%% file: arxiv.tex
\documentclass{article}
\usepackage[preprint]{neurips_2024}
\usepackage{xcolor}
\usepackage[utf8]{inputenc} 
\usepackage[T1]{fontenc}    
\usepackage{hyperref}       
\usepackage{url}            
\usepackage{booktabs}       
\usepackage{amsfonts}       
\usepackage{nicefrac}       
\usepackage{microtype}      
\usepackage{amsmath}
\usepackage{xspace}
\usepackage{amssymb}
\usepackage{graphicx}
\usepackage{multirow}
\usepackage{enumitem}
\usepackage{ulem}
\usepackage{float}
\usepackage{makecell}
\usepackage{caption}
\usepackage{threeparttable}
\usepackage{listings}
\usepackage[frozencache,cachedir=.]{minted}

\newcommand{\ie}{\textit{i.e.,}\xspace}
\newcommand{\eg}{\textit{e.g.,}\xspace}
\newcommand{\etal}{\textit{et al.}\xspace}
\newcommand{\paratitle}[1]{\vspace{0.8ex}\noindent \textbf{#1}}

\newcommand{\archiname}{Sketch\xspace}

\definecolor{bggray}{RGB}{245, 245, 245}

\title{\archiname: A Toolkit for Streamlining LLM Operations}

\author{
  Xin Jiang\textsuperscript{1\textdagger},
  Xiang Li\textsuperscript{1\textdagger}, 
  Wenjia Ma\textsuperscript{2\textdagger}, 
  Xuezhi Fang\textsuperscript{1}, 
  Yiqun Yao\textsuperscript{1}, 
  Naitong Yu\textsuperscript{1},\\
  \textbf{Xuying Meng\textsuperscript{3}, Peng Han\textsuperscript{4}, Jing Li\textsuperscript{5}, Aixin Sun\textsuperscript{6}, Yequan Wang\textsuperscript{1$*$}}\\
  $^{1}$Beijing Academy of Artificial Intelligence, Beijing, China\\
  $^{2}$AstralForge AI Lab, Beijing, China\\
  $^{3}$Institute of Computing Technology, Chinese Academy of Sciences, Beijing, China\\
  $^{4}$University of Electronic Science and Technology of China, Chengdu, China\\
  $^{5}$Harbin Institute of Technology, Shenzhen, China\\
  $^{6}$College of Computing and Data Science, Nanyang Technological University, Singapore
}

\begin{document}

\maketitle

\renewcommand{\thefootnote}{\fnsymbol{footnote}}
\footnotetext[2]{Indicates equal contribution.}
\footnotetext[1]{Corresponding author.}
\renewcommand{\thefootnote}{\arabic{footnote}}

\begin{abstract}
Large language models~(LLMs) represented by GPT family have achieved remarkable success. The characteristics of LLMs lie in their ability to accommodate a wide range of tasks through a generative approach. However, the flexibility of their output format poses challenges in controlling and harnessing the model's outputs, thereby constraining the application of LLMs in various domains. In this work, we present \archiname, an innovative toolkit designed to streamline LLM operations across diverse fields. \archiname comprises the following components: (1) a suite of task description schemas and prompt templates encompassing various NLP tasks; (2) a user-friendly, interactive process for building structured output LLM services tailored to various NLP tasks; (3) an open-source dataset for output format control, along with tools for dataset construction; and (4) an open-source model based on LLaMA3-8B-Instruct that adeptly comprehends and adheres to output formatting instructions. We anticipate this initiative to bring considerable convenience to LLM users, achieving the goal of ``plug-and-play'' for various applications. The components of \archiname will be progressively open-sourced at \url{https://github.com/cofe-ai/Sketch}.

\end{abstract}

\section{Introduction}
\label{sec:intro}

Generative pre-trained large language models~(LLMs) have achieved remarkable success, with notable examples including GPT~\cite{gpt3,GPT-4}, LLaMA~\cite{llama1,llama-2,Llama3}, and FLM~\cite{freelm,flm101b,teleflm} series. One of the key advantages of these models lies in their powerful generalization capabilities: a single model is capable of handling a diverse range of tasks.
However, accurately generating formatted outputs, such as JSON, remains challenging for LLMs because they do not always strictly follow instructions.
On the demand side, AI-driven applications urgently require the integration of structured outputs (\eg JSON) from LLMs into their data streams. This has heightened the urgency for LLMs to produce controlled and structured outputs as demanded.

The requirement for structured outputs from LLMs can be resolved through a multitude of approaches.
In-context learning is a typical approach. It not only enhances model performance but also offers a certain degree of format control without incurring additional computational costs for model fine-tuning.
However, this approach faces challenges, such as an inability to determine when to end the generation. 
Besides, it needs long-text ability when meeting complex questions, as it relies on extensive input examples to ensure accurate decision-making. 
Moreover, tasks that require complex constraints on format and content, such as relation extraction and event extraction, pose significant difficulties for in-context learning.

Supervised fine-tuning~(SFT) refers to the process of training a pre-trained model on a labelled dataset specifically tailored for a particular task.
Although SFT can enhance performance on specific tasks and has generalization capabilities, its ability to control the format of the output remains unsatisfactory.
After all, the integration of LLM outputs into applications typically demands the output format that is entirely compliant with specified requirements, a feat that LLMs, proficient in ``next token prediction'', are unable to ensure.
Another issue is that, to the best of our knowledge, there is a lack of open-source models and datasets specifically addressing the problem of formatted output control.
This somewhat limits the application of LLMs across various fields.

To ensure that the outputs of LLMs conform to formatting requirements, numerous decoding control tools~(guidance\footnote{\url{https://github.com/guidance-ai/guidance}}, outlines\cite{willard2023efficient}, llama.cpp\footnote{\url{https://github.com/ggerganov/llama.cpp}}, lm-format-enforcer\footnote{\url{https://github.com/noamgat/lm-format-enforcer}} ) based on regular expressions or context-free grammars~(CFGs) have been developed. 
These tools first convert the user's requirements for output format into formal languages. Under the constraints of these formal languages, these models could decode responses that meet the formatting requirements. More importantly, as these tools are involved in the decoding process of the model, they could potentially impair the model's performance\cite{tam2024let}, especially if the model itself is not adept at generating structured outputs.
To address those issues, an open-source model that excels in generating structured responses according to requirements, along with a framework for streamlining various LLM-based operations, holds significant value.

In this work, we introduce \archiname, a toolkit designed to assist users in effectively operating LLMs and generating results in their expected format.
The core idea of \archiname is as follows: targeting on various NLP tasks, we establish a collection of task description schema, within which users can delineate their own tasks, including task objectives, labelling systems, and most critically, the specifications for the output format. An LLM can then be deployed out of the box to handle these unfamiliar tasks, ensuring the correctness and conformity of the output format. This approach not only streamlines the process for users but also enhances the reliability and precision of the model's outputs, making it a versatile and robust solution for a wide array of NLP applications.

The main contributions are as follows:
\begin{itemize}
    \item We propose \archiname, an innovative operating framework simplifying the process for LLM users, enabling ``plug-and-play'' functionality for task-specific applications with predefined schemas. The proposed \archiname makes it easier to instantiate and manage NLP tasks.
    
    \item To optimize the performance within \archiname framework, we build a dataset and conduct model fine-tuning based on LLaMA3-8B-Instruct, ensuring superior task handling and output consistency. 
    Both the dataset and fine-tuned model will soon be made available to the public.

    \item By integrating constrained decoding frameworks, \archiname ensures precise control over the model's output format, enhancing the reliability and precision of outcomes, and facilitating direct application of large models in industry settings.
\end{itemize}

\section{\archiname Architecture}
\label{sec:schema}

\begin{figure}
    \centering
    \includegraphics[width=1.0\linewidth]{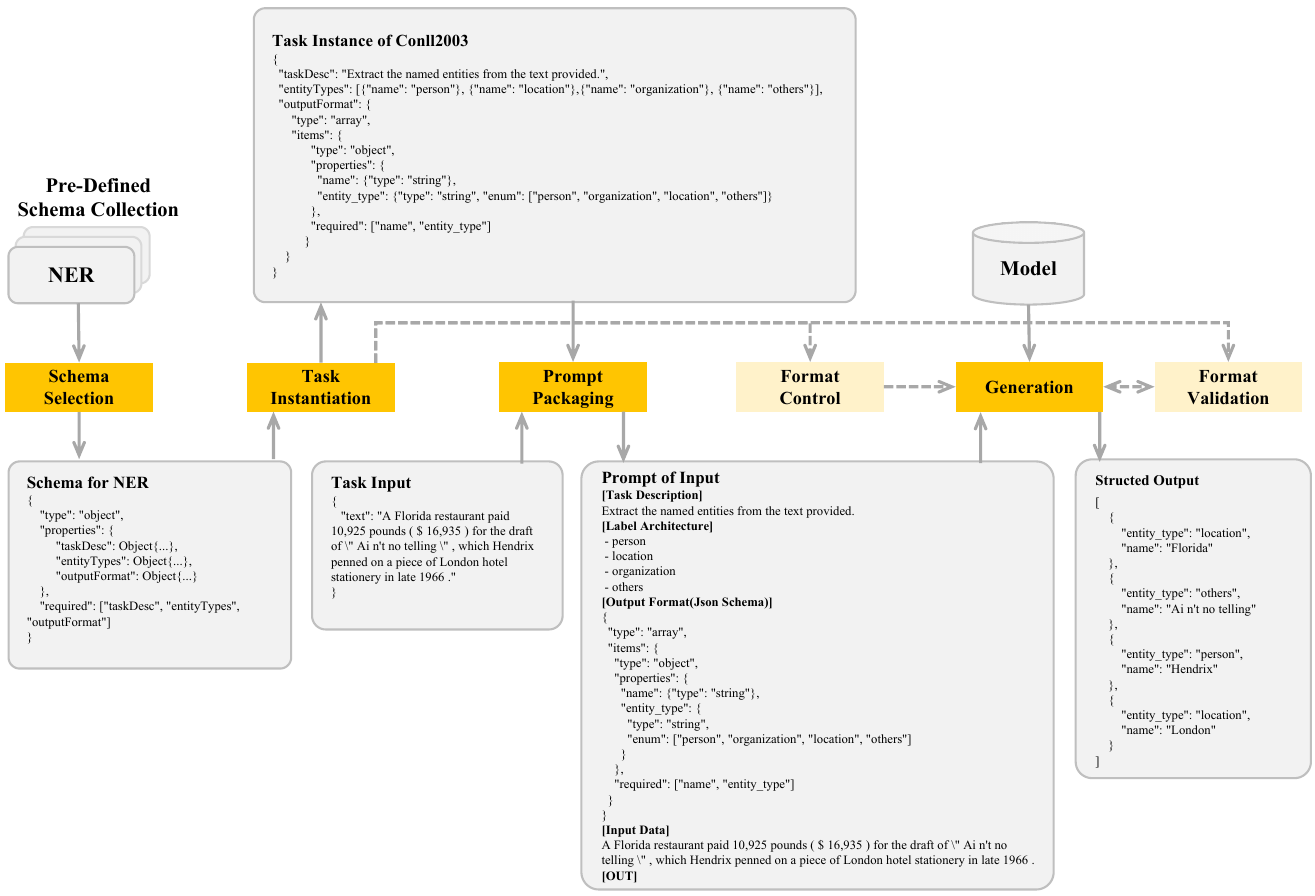}
    \caption{\archiname workflow, taking a NER task CoNLL-2003 as an example. The nodes (Format Control and Format Validation) in slight yellow are optional.}
    \label{fig:workflow}
\end{figure}

\archiname is designed to enable controlled formatting and easy interaction with LLMs. In this section, we detail the architecture of \archiname and how to use it easily. Figure~\ref{fig:workflow} illustrates the concepts and internal workflow of \archiname. The workflow consists of four steps: \textit{schema selection}, \textit{task instantiation}, \textit{prompt packaging}, and \textit{generation}. In practical applications, the complex aspects of this process are transparent to the user.

First, users are guided to choose the appropriate schema from a predefined set that aligns with the specific NLP task requirements. A schema, in essence, is a class (or a JSON Schema~\footnote{\url{https://json-schema.org}} in practice) that standardizes the user's description of tasks.
Second, in the task instantiation phase, users populate the chosen schema with task-specific details such as description, label set, choice type, and output format, resulting in a task instance(in JSON format) that adheres to the corresponding schema.
Third, based on the task instance, the prompt packaging step involves \archiname automatically converting the task input into a structured prompt tailored for LLM interaction.
Last, during generation, \archiname can not only infer the model to get the anticipated response but also optionally integrate the lm-format-enforcer, a control architecture that constrains LLM outputs to comply with the specified output format.

\subsection{Schema Selection}

Schema is the bridge between tasks and LLMs. 
It outlines a descriptive framework for each kind of task based on the task-specific characteristics.
A schema can be represented by either a Pydantic model or a JSON Schema.
When customizing a specific task, users are advised to select the most appropriate schema and instantiate the task within its constraints.
This process can be achieved through a Python API, and we also provide a more intuitive interactive method in the form of filling out a form generated by \archiname based on the schema.
To date, as the initial phase of \archiname's development, we have experimentally built a set of schemas for tasks, including over ten subcategories under the three main categories of text classification, text generation, and information extraction, as shown in Table\ref{tab:schema-design}.
A selection of the schemas we have crafted is showcased in Appendix~\ref{sec:schema-example}. 
For an extensive view of the task schemas available, please visit our project repository at \url{https://github.com/cofe-ai/Sketch}
\input{tables/schema-design}

\subsection{Task Instantiation}
We define \textit{Task Instance} as a standardized description of a particular task within the constraints of the schema it belongs to, and the process of creating it by the user is referred to as \textit{Task Instantiation}. A task instance typically includes the following basic fields:

\textit{Task specification fields} delineates the task, which may encompass the ``taskDesc'' field detailing the task's purpose, along with the ``labelSet'' and ``choiceType'' fields that respectively define the classification schema and the number of options. We establish different required fields for various tasks.

\textit{Output format field} specifies and constrains the format that users expect the model to output. We choose JSON schema as the descriptive language. This field serves a dual function: it is integrated into the prompt to direct the model's output, and it is also converted into a decoding control mechanism in the form of regular expressions, typically enforced through finite state machines (FSMs). This strategy intervenes in the model's decoding process to guarantee that the output is 100\% compliant in form with the users' expectations.

A comprehensive illustration of task instances, replete with intricate details and concrete examples, can be found in Appendix \ref{sec:ti-example}.

\subsection{Prompt Packaging}
The process of packaging an instantiated task and input is crucial for ensuring LLMs understand the task requirements and process the input correctly. This step involves combining the structured task description with the user's input data into a format that is optimized for interaction with the LLM.

\paratitle{Input Integration.} The user's input, whether it be a common text snippet or any other form of information relevant to the task, will be integrated into the prompt most intuitively. This integration is guided by a prompt template associated with the schema, ensuring that the input is presented in a manner that is coherent and comprehensible to LLMs.

As shown in Figure~\ref{fig:workflow}, for a NER task, the packaged prompt might include \textit{[Task Description]}, \textit{[Label Architecture]}, \textit{[Output Format~(Json Schema)]}, and \textit{[Input Data]} to be processed. This ensures that the LLM understands the task criteria and outputs the result in the desired format.

\subsection{Generation}

The final step in the workflow of \archiname involves the interaction with LLMs to generate the desired output.
\archiname is able to generate the expected response directly with a good performance. Besides, there are some more precise control methods. Throughout this process, we ensure that the model's output conforms to the required format from two perspectives.

\textbf{Constrained Generation.} Considering that even with meticulous fine-tuning, LLMs cannot guarantee 100\% accuracy in output format, we integrate a mature decoding control framework, lm-format-enforcer. It employs CFG to ensure that the model’s responses align perfectly with the predefined output format.
Simultaneously, recognizing that any constraints to the decoding process may impact the model's performance, this strict output format control is made optional in \archiname.

\textbf{Output Validation.} Given that not all JSON Schema properties are supported by decoding control frameworks, the output produced by the LLM cannot be assured to adhere to the constraints of the specified output format. To ensure compliance with the expected format, we employ the \textit{jsonschema} tool\footnote{\url{https://github.com/python-jsonschema/jsonschema}} for validation. For outputs that do not meet the expected format, we take measures such as resampling or directly throwing exceptions. 

By following these detailed steps, \archiname ensures that users can effectively utilize LLMs for a variety of NLP tasks, with the assurance that the outputs will be both accurate and in the correct format. This streamlined process makes it easier for users to interact with complex models and harness their power for practical applications.

\subsection{Code Example}
Listing \ref{lst:python_code} demonstrates the basic usage of \archiname through a simple named entity recognition (NER) task. \archiname is still under development prior to its release, and the APIs may change at any time.

\input{tables/code-example}

\section{\archiname-8B Fine-tuning Approach}
\label{sec:model}
We fine-tune LLaMA3-8B-Instruct to enhance the model's capability to generate structured data that adheres to JSON schema constraints across a variety of tasks. Our training process emphasizes two key aspects: ensuring strict adherence to the specified JSON schema constraints in the model's outputs and fostering robust generalization across various tasks. To achieve these goals, we carefully design a specialized fine-tuning dataset.

\subsection{Data Preparation}
The capability of the model to adhere to formats and its ability to understand and tackle tasks are distinct attributes. To enhance these aspects, we have constructed two targeted datasets: NLP task data and schema following data. The primary objective of NLP task data is to enable models to learn how to tackle NLP tasks. However, considering the limitations in output format diversity of manually curated fine-tuning data for NLP tasks, we propose the automated construction of schema following data to enhance the model's adherence to the output format schema.

\textbf{NLP Task Data.}
We assemble a comprehensive collection of over 20 datasets, encompassing more than ten subcategories within three primary domains: text classification, text generation, and information extraction. Through the meticulous design of output formats for each dataset, we construct a task instance set of size 53. Among them, 37 task instances are dedicated to training, while the remainder are reserved for evaluation.

\textbf{Schema Following Data.}
To ensure the diversity of JSON schemas, we generated 10,000 JSON schemas with widths and depths within 5 with a random schema generation method. Then, we utilized LLaMA3-8B-Instruct, under the constraint of a decoding control tool, to generate JSON instances that conform to the schemas. 
Following the patterns of NLP task data, we designed a task that involves selecting values from a randomly generated list of given values to construct JSON objects that match specific schemas.
Finally, we constructed 20,000 pieces of fine-tuning data for this task by modifying the values in the JSON instances generated by LLaMA3-8B-Instruct.

\subsection{Fine-tuning Method}

Reinforcement learning is one of the popular ways to tune the LLMs. LeCun holds the opinion, ``I do favor MPC over RL''\footnote{\url{https://x.com/ylecun/status/1827787323108393027}}. We have a similar opinion so we use the easy fine-tuning methods under data-driven. Indeed, it doesn't mean reinforcement learning is useless, but it could be used in the following steps such as resort. Generating valid outputs that conform to the JSON Schema is not simply a matter of mimicking formats, it necessitates a thorough comprehension of the schema's descriptions. Consequently, data adhering to the schema is essential for enhancing the model's ability.

The training objective of \archiname-8B considers two aspects: enhancing the model's adherence to format and improving its NLP task performance. To this end, we use the proposed mixed dataset comprising NLP task data and schema following data for fine-tuning. 
The inclusion of NLP task data markedly boosts the model’s capabilities in handling NLP tasks while the schema following data is crucial for enhancing the model’s adherence to various output format requirements.

We use fine-tuning method to optimize the proposed model, the objective $\mathcal{L}(\theta)$ could be formatted as:
\begin{equation}
    \mathcal{L}(\theta) = -\sum^{m}_{t=1} log{P_{\theta}(\hat{y_t}=y_t|y_{1:t-1},X)}
\end{equation}
where $X = \{x_1, x_2, \dots, x_n\}$ represents an input sequence of length $n$, which is the constructed prompt. $Y = \{y_1, y_2, \dots, y_m\}$ is the label of the generated sequence of length $m$, and $\hat{Y} = \{\hat{y}_1, \hat{y}_2, \dots, \hat{y}_m\}$ is the actual output of the model. Note that both $Y$ and $\hat{Y}$ exclude the prompt and consist only of the response. $\theta$ denotes the model parameters, and $P_{\theta}$ represents the conditional probability under the parameters $\theta$.

Each sample consisting of $X$ and $Y$ is sampled from a carefully constructed mixed dataset. The optimal fine-tuning effect is achieved by appropriately balancing the ratio of NLP task data to schema-following data in the mixed dataset.

\section{Experiments}
\label{sec:exp}

\subsection{Experiment Settings}

In this section, we validate the model's generalization capabilities through experiments and discuss the effectiveness and optimal configuration of our fine-tuning data. 

\paragraph{Experiment Data Settings.}
We use publicly available NLP task datasets~(See Appendix~\ref{sec:datasets} for details) for experiments. For each dataset, we carefully construct different task instances, expanding a single dataset into multiple experimental datasets with varying \textit{outputFormat} and other task-related parameter configurations. To validate different hypotheses, we selectively exclude some data from the training set to create test datasets. These test datasets include three types: (1) the output format not seen in the training set while other output formats from the same dataset are included, (2) the entire dataset is not present in the training set, and (3) the entire tasks are not included in the training set.

\paragraph{Fine-tuning Settings.}
We experiment on LLaMA3-8B-Instruct since it has strong foundational capabilities. We fine-tune the model for 8 epochs with a global batch size of 128, setting the learning rate to 1e-6 and weight decay to 0.1. The learning rate is decayed to 0 using a linear schedule. We select the best checkpoint from the model at the end of every epoch.

\paragraph{Evaluation Methods.}
To comprehensively evaluate the model's schema adherence and NLP task performance, we assess from two perspectives:
\begin{enumerate}
    \item We define a metric to assess the model's ability to produce outputs that conform to the \textit{outputFormat}: Legal Output Ratio. First, we determine whether the model's output can be converted into a JSON object; if not, the output is considered invalid. Next, we check if the JSON object meets the \textit{outputFormat} requirements; otherwise, it is considered invalid. The legal output ratio is calculated by dividing the number of valid outputs by the total number of test samples.
    \item To evaluate NLP task performance, we employ traditional metrics like F1-score or accuracy, tailored to the specific requirements of each task.

\end{enumerate}

\subsection{Comparison with Baselines}
To evaluate generalization, we fine-tune \archiname-8B-w.o.-ner with a partially removed dataset and benchmark it against mainstream models, including GPT-4o, DeepSeek, and ChatGLM. Using identical prompts across models, we gather results via API and assess performance. We also compare \archiname-8B-w.o.-ner with the original LLaMA3-8B-Instruct (local inference). Additionally, we evaluate DeepSeek's one-shot results and GPT-4o's constrained decoding. \archiname-8B-w.o.-ner and LLaMA-8B-Instruct use FSM and CFG constraints for decoding. The comparison covers three dataset types: (1) \textbf{unknown format}, with output formats absent in training data, (2) \textbf{unknown domain}, with datasets from untrained domains, and (3) \textbf{unknown task}, focusing on task types not covered during training. NER is the test task for the Unknown Task category.

\input{tables/exp-main}

\paratitle{Schema Adherence Comparison.}
Table~\ref{tab:exp-main} illustrates notable differences in schema adherence among baseline models under unconstrained output conditions. For simpler formats like S10T8 and HOTEL, LLaMA3-8B-Instruct achieves nearly 100\% on legal output ratio but fails completely on 20NEWS. Across most datasets, its legal output ratio ranges from 50\% to 75\%, averaging 64.9\%. In contrast, \archiname-8B-w.o.-ner achieves an average legal output ratio of 96.2\% under unconstrained conditions, with its lowest performance on CNL03 still at 83.8\%. This demonstrates \archiname-8B-w.o.-ner's strong generalization in format adherence.

\paratitle{Performance Comparison.}
We compare with LLaMA3-8B-Instruct to assess training effectiveness and with mainstream models to evaluate performance level:

1. \textbf{vs LLaMA3-8B-Instruct}. Table~\ref{tab:exp-main} shows that \archiname-8B-w.o.-ner consistently outperforms LLaMA3-8B-Instruct under the same decoding strategy, both on individual subsets and in average scores. Furthermore, the unconstrained \archiname-8B-w.o.-ner surpasses LLaMA3-8B-Instruct across all decoding strategies. The results indicate that the fine-tuning method enhances NLP task performance and demonstrates strong generalization to unknown output formats and tasks.

2. \textbf{vs Mainstream Models}. Comparing \archiname-8B-w.o.-ner with mainstream models like DeepSeek, ChatGLM, and GPT-4o on unknown format datasets, \archiname-8B-w.o.-ner significantly outperforms all, achieving nearly 100\% legal output ratio on 20NEWS where others struggle (\eg GPT-4o below 50\%). On unknown domain datasets, it performs similarly to DeepSeek and GPT-4o but surpasses ChatGLM. However, its smaller model size leads to some limitations on unknown task datasets compared to larger models.

\paratitle{Constrained Decoding Evaluation.}
The analysis also reveals that FSM and CFG constraints do not consistently produce the expected outcomes. FSM constraints result in lower task evaluation scores for both \archiname-8B and LLaMA3-8B-Instruct. While CFG constraints improve overall average scores, they fail to enhance task evaluation scores on datasets with hard output formats (\eg 20NEWS), despite increasing the legal output ratio. This suggests that current constrained decoding methods are not yet consistently reliable for real-world NLP tasks.

\subsection{Generalization Capability Analysis}

\paragraph{Output Format Generalization Capability.}

We first evaluate \archiname-8B's generalization capability across different output formats within the same dataset.
As shown in the ``Unknown Format'' column of Table~\ref{tab:exp-main}, the output formats of the two datasets (S10T8 and 20NEWS) used for evaluation are not in \archiname-8B's training set.
We can observe that in S10T8, both LLaMA3-8B-Instruct and \archiname-8B achieve high precision (0.997 and 0.982) in adhering to the required output format, which is likely due to the format simplicity. 
For 20NEWS, due to the complex format, LLaMA3-8B-Instruct is completely unable to follow the required output format. Surprisingly, despite not being trained in this specific format, \archiname-8B shows an impressive ability to follow output format. This demonstrates \archiname-8B's generalization ability on unseen formats within the trained dataset.

\paragraph{Domain Generalization Capability.}

Further, we evaluate \archiname-8B's cross-domain generalization capability within the same task (\eg NER). This is crucial for models' application in various scenarios from various users. We continue to evaluate \archiname-8B on two tasks: aspect-level sentiment analysis, and text topic classification. We construct two datasets that are untrained and completely different from the domains of \archiname-8B's training datasets. The results in Table \ref{tab:exp-main} column ``Unknown Domain'' show that \archiname-8B significantly outperforms LLaMA3-8B-Instruct on these two datasets~(domains), both in terms of adherence to formatting requirements and NLP task F1/Accuracy. It is important to note that \archiname-8B has never encountered data from these three domains during training. 
This illustrates a fact: \archiname-8B is capable of enhancing its performance across different domains within a task by training on data associated with specific domains (or taxonomies). 

\paragraph{Task Generalization Capability.}
Ultimately, we evaluate the ability to generalize across tasks. This ability is known as the most formidable aspect of generalization. While we can endeavour to build an extensive array of NLP task categories, the spectrum of potential tasks is infinite. As such, LLM users across a myriad of sectors undoubtedly desire a model that can extend its reach to cover their unique and unconventional task needs. This is why we present the evaluation results of \archiname-8B-w.o.-ner in Table~\ref{tab:exp-main}. We completely exclude the NER datasets from the training set and evaluate how the output format following capabilities of \archiname-8B-w.o.-ner improve on NER tasks.
Remarkably, \archiname-8B-w.o.-ner demonstrates significant improvement in the two NER datasets, with L.O.R. increasing from 0.520 to 0.939 and from 0.645 to 0.968, respectively.
Consequently, it can be concluded that for an unfamiliar NLP task, \archiname-8B is likely a superior choice compared to LLaMA3-8B-Instruct, even though it has not been trained on such tasks.

\subsection{Data Configuration Experiment}
\label{sec:exp_data_config}
Fine-tuning data is central to this work. We analyze how data proportion and scale affect model performance. The evaluation focuses on the model's results on a test set with seven tasks: three with unseen output formats, three from unseen domains, and one entirely new task.

\input{tables/exp-data-weight}

\paratitle{Data Proportion.}
Different sampling proportion affects the performance of pretraining foundation models. Similar to this phenomenon, the sampling proportion of schema following data leads to a decline in task performance.

To assess the effectiveness and configuration of NLP task data and schema following data, we conduct experiments using a fixed 20k dataset with various proportions, including a setup without schema following data. Performance is evaluated on the test set, with results shown in Table \ref{tab:exp-data-weight}.

From the table, we observe that the schema following data proportion is positively correlated with the legal output ratio. Schema following data significantly enhances the model's ability to follow output formats. However, when the schema following data proportion exceeds 25\%, performance on the test set declines from 0.688 to 0.655, indicating that excessive schema following data negatively impacts task performance. Therefore, we determine that a 7:1 ratio of Task Data to schema following data is optimal.

\input{tables/exp-data-size}

\paratitle{Data Volume.}
We conduct experiments to analyze the impact of data size on results. Four fine-tuning datasets with 10k, 20k, 30k, and 40k samples (with a 7:1 ratio of Task Data to schema following data) are used to train the model, which are then evaluated on the same test set.

As shown in Table \ref{tab:exp-data-size}, the best legal output ratio 0.697 is achieved with the 30k dataset. Increasing the data size to 40k leads to a noticeable decline in both performance and legal output ratio. This suggests that more fine-tuning data does not always yield better results. We ultimately select the 30k dataset for training the \archiname-8B model.

\section{Related Work}
\label{sec:related}

Significant advancements have been made in the realm of format-constrained generation for LLMs. We roughly divide these methods into three categories: pre-generation tuning, in-generation control, and post-generation parsing.

\paragraph{Pre-generation Tuning.}
Pre-generation tuning encompasses a suite of techniques designed to fine-tune the behaviour of LLMs before the actual text-generation process begins. This approach involves modifying the model's training data\cite{zhou2023controlledtextgenerationnatural,yao2024opendomainimplicitformatcontrol} or prompts\cite{DBLP:conf/nips/BrownMRSKDNSSAA20,DBLP:conf/nips/Wei0SBIXCLZ22} to align more closely with the specific format constraints required by the task at hand.

\paragraph{In-generation Control.}

There are numerous frameworks dedicated to intervening in the decoding process of LLMs to control the permissible range of the next token, ensuring that the output of the LLM meets the format requirements. The predominant control strategies employed include JSON Schema~(\ie Jsonformer\footnote{\url{https://github.com/1rgs/jsonformer}}, lm-format-enforcer and outlines), regular expression~(\ie guidance, lm-format-enforcer
 and outlines) and context-free grammar
~(\ie llama.cpp). Although these methods typically ensure high accuracy in response format, they often lead to a decrease in the usefulness of the responses\cite{tam2024let}, which is one of the starting points for the work presented in this paper.

\paragraph{Post-generation Parsing.}
This category involves techniques that parse the output of LLMs after generation to ensure it conforms to specific formats. These methods often rely on post-processing algorithms to refine the raw output into a structured format. Guardrails\footnote{\url{https://github.com/guardrails-ai/guardrails}} is a framework of this kind, designed to enforce constraints on the output of LLMs by filtering or modifying the generated text to ensure it adheres to predefined guidelines or specifications.

\section{Conclusions and Future Work}
\label{sec:con}

In this work, we propose \archiname to simplify and optimize the applications of LLMs. Using a schema-based approach, \archiname can tackle the challenges in structured output generation and model generalization. Key contributions include the schema architecture for task description, data and model fine-tuning for improved performance, and the integration of a constrained decoding framework for precise output management. 
Experimental results not only demonstrate the enhanced capability of the fine-tuned \archiname-8B in adhering to output formats but also validate the effectiveness of the fine-tuning data we build, particularly the schema following data.

Future work involves expanding task categories, optimizing model performance, lowering entry barriers, and exploring new applications in diverse domains. \archiname's innovative approach and ongoing development promise to drive advancements in LLM applications and unlock new possibilities for harnessing the power of LLMs.

\section*{Acknowledgments}
This work is supported by the National Science and Technology Major Project (No. 2022ZD0116300) and the National Science Foundation of China (No. 62106249).

\newpage
\appendix

\section{Schema Examples}
\label{sec:schema-example}
\input{tables/schema-example-NER}
\input{tables/schema-example-TC}
\input{tables/schema-example-Translation}

\newpage
\section{Task Instance Examples}
\label{sec:ti-example}
\input{tables/task-instantiation-NER}
\input{tables/task-instantiation-TC}
\input{tables/task-instantiation-Translation}

\newpage
\section{NLP Task Datasets}
\label{sec:datasets}
In this appendix, we provide a detailed description of the datasets utilized in this paper. These datasets are categorized into one of the following task categories:
\begin{itemize}[leftmargin=0.5cm]
\item \textbf{Information extraction} Information extraction (IE) encompasses the task of discerning and extracting structured information from unstructured and/or semi-structured machine-readable documents. This category includes various sub-tasks such as relation extraction, named entity recognition, event extraction, and aspect-level sentiment analysis. The following datasets are utilized to facilitate these tasks: 
\begin{enumerate}
    \item \textbf{Relation extraction}: SemEval-2010 Task 8\cite{DBLP:journals/corr/abs-1911-10422}, TACRED\cite{DBLP:conf/emnlp/ZhangZCAM17};
    \item \textbf{Named entity recognition}: CoNLL-2003\cite{DBLP:conf/conll/SangM03}, UniversalNER\cite{DBLP:conf/iclr/Zhou00CP24}, Medical NER\cite{corona};
    \item \textbf{Aspect-level sentiment analysis}: SemEval-2014 Task 4\cite{DBLP:conf/semeval/PontikiGPPAM14}; SemEval-2015 Task 12\cite{DBLP:conf/semeval/PontikiGPMA15}; 
    \item \textbf{Event extraction}: DuEE\cite{DBLP:conf/nlpcc/LiLPCPWLZ20};
\end{enumerate}

\item \textbf{Text classification} Text classification is the task of assigning predefined categories to text documents. It encompasses a wide range of tasks, such as sentiment analysis, topic classification, intent recognition, sentence similarity, \etal We use the following datasets:
\begin{enumerate}
    \item \textbf{Sentiment analysis}: APP\_REVIEWS\cite{DBLP:conf/sigsoft/GranoSMVCP17}, ChnSentiCorp\cite{DBLP:journals/eswa/TanZ08}, IMDB\cite{maas-EtAl:2011:ACL-HLT2011};
    \item \textbf{Topic classification}: 20 Newsgroups\cite{LANG1995331}, AG News\cite{amazon_polarity}, BBC News\cite{DBLP:conf/icml/GreeneC06}, DBPedia\cite{amazon_polarity};
    \item \textbf{Intent recognition}: MASSIVE\cite{DBLP:conf/acl/FitzGeraldHPMRS23}, BANKING77\cite{DBLP:journals/corr/abs-2003-04807};
    \item \textbf{Sentence similarity(also known as paraphrase detection)}: QQP\cite{DBLP:conf/iclr/WangSMHLB19};
    \item \textbf{Natural language inference}: RTE\cite{rte1};
\end{enumerate}

\item \textbf{Text generation} Text generation involves creating text from scratch or completing partial texts based on given prompts. This task is essential for applications such as chatbots, translation, and summarization. We use the following datasets:
\begin{enumerate}
    \item \textbf{Summarization}: xsum\cite{DBLP:conf/emnlp/NarayanCL18};
    \item \textbf{Translation}: Replete-AI/Multi-lingual$\_$Translation$\_$Instruct\footnote{\url{https://huggingface.co/datasets/Replete-AI/Multi-lingual_Translation_Instruct}};
    \item \textbf{Dialog}: shibing624/sharegpt$\_$gpt4\footnote{\url{https://huggingface.co/datasets/shibing624/sharegpt_gpt4}};
\end{enumerate}

\end{itemize}

\end{document}

%% file: tables/schema-design.tex
\begin{table}[htbp]
  \centering
  \caption{The \archiname architecture of LLMs for tasks}
  \scalebox{0.85}{
    \begin{tabular}{cll}
    \toprule
    Category & Task  & Required fields \\
    \midrule
    \multirow{5}[2]{*}{Text classification} & Topic classification  & \multirow{5}[2]{*}{\textit{taskDesc, labelSet, choiceType, outputFormat}} \\
          & Sentiment analysis &  \\
          & Sentence similarity &  \\
          & Intent recognition &  \\
          & Natural language inference &  \\
    \midrule
    \multirow{3}[2]{*}{Text generation} & Summarization & \multirow{3}[2]{*}{\textit{taskDesc, outputFormat }} \\
          & Dialog &  \\
          & Translation &  \\
    \midrule
    \multirow{5}[2]{*}{Information extraction} & Relation Extraction & \textit{taskDesc, relationTypes, outputFormat} \\
          & Named entity recognition  & \textit{taskDesc, entityTypes, outputFormat} \\
          & Keyword extraction  & \textit{taskDesc, outputFormat} \\
          & Event extraction  & \textit{taskDesc, eventTypes, outputFormat} \\
          & Aspect-level sentiment analysis  & \textit{taskDesc, sentimentTypes, outputFormat} \\
    \bottomrule
    \end{tabular}%
    }
  \label{tab:schema-design}%
\end{table}%

%% file: tables/code-example.tex
\setminted{
    frame=lines,
    framesep=2mm,
    baselinestretch=1.2,
    fontsize=\footnotesize
}

\begin{minipage}{\linewidth}
\captionof{listing}{Example of \archiname's Usage} 
\label{lst:python_code} 
\begin{minted}[autogobble,frame=none, bgcolor=bggray]{python}
import llm_sketch
from transformers import AutoModelForCausalLM, AutoTokenizer

model = AutoModelForCausalLM.from_pretrained("CofeAI/Sketch-8B", device_map="auto")
tokenizer = AutoTokenizer.from_pretrained("CofeAI/Sketch-8B")

my_ner_task = llm_sketch.schemas.NER(
    taskDesc="Extract the named entities from the given text.",
    entityTypes=[
        {"name": "person"}, 
        {"name": "organization"}, 
        {"name": "location"}
    ],
    outputFormat={
        "type": "array",
        "items": {
            "type": "object",
            "properties": {
                "name": {"type": "string", "description": "the entity name"},
                "entity_type": {
                    "type": "string",
                    "description": "entity type",
                    "enum": ["person", "organization", "location"],
                },
            },
            "required": ["name", "entity_type"],
        },
    },
)

inputs = [
    "Kamala Harris pledges 'new way forward' in historic convention speech"
]
for inpt in inputs:
    ner_res = llm_sketch.generate(model, tokenizer, inpt, my_ner_task, strict=True)
    print(ner_res)
# [{'name': 'Kamala Harris', 'entity_type': 'person'}]

\end{minted}
\end{minipage}

%% file: tables/exp-main.tex
\begin{table*}
  \centering
  \caption{\textbf{Evaluation with format output.} L.O.R. (Legal Output Rate) is the proportion of model outputs correctly formatted according to the JSON schema. The datasets S10T8, 20NEWS, HOTEL, DBP14, CNL03, and MED correspond to: SemEval-2010 Task 8, 20 Newsgroup, SemEval-2015 Task 12 (domain: hotels), DBPedia, CoNLL-2003, and Medical NER. The task types RE, CLS, ASA, and NER stand for Relation Extraction, Topic Classification, Aspect-Level Sentiment Analysis, and Named Entity Recognition.}
  \small{
    \begin{tabular}{cl|cc|cc|cc|c}
    \toprule
    \multirow{3}[2]{*}{Models} & \multicolumn{1}{c|}{\multirow{3}[2]{*}{Index}} & \multicolumn{2}{c|}{Unknown Format} & \multicolumn{2}{c|}{Unknown Domain} & \multicolumn{2}{c|}{Unknown Task} & \multirow{3}[2]{*}{Avg} \\
          &       & S10T8 & 20NEWS & HOTEL & DBP14 & CNL03 & MED   &  \\
          &       & RE    & CLS   & ASA   & CLS   & NER   & NER   &  \\
    \midrule
    \multirow{2}[2]{*}{DeepSeek} & L.O.R. & 1.0   & 0.0   & 1.0   & 1.0   & 1.0   & 1.0   & 0.833 \\
          & F1/Acc. & 0.270 & 0.0   & 0.500 & 0.890 & 0.647 & 0.583 & 0.482 \\
    \midrule
    DeepSeek & L.O.R. & 1.0   & 0.040 & 0.020 & 1.0   & 1.0   & 0.0   & 0.510 \\
    (one-shot) & F1/Acc. & 0.380 & 0.040 & 0.0   & 0.930 & 0.641 & 0.0   & 0.332 \\
    \midrule
    \multirow{2}[2]{*}{ChatGLM} & L.O.R. & 0.970 & 0.030 & 0.990 & 0.980 & 0.860 & 1.0   & 0.805 \\
          & F1/Acc. & 0.386 & 0.020 & 0.367 & 0.930 & 0.671 & 0.554 & 0.488 \\
    \midrule
    \multirow{2}[2]{*}{GPT-4o} & L.O.R. & 1.0   & 0.460 & 1.0   & 1.0   & 1.0   & 1.0   & 0.910 \\
          & F1/Acc. & 0.510 & 0.340 & 0.551 & 0.900 & 0.716 & 0.618 & 0.606 \\
    \midrule
    GPT-4o & L.O.R. & 1.0   & -     & -     & 1.0   & 1.0   & -     & - \\
    (constrained with CFG) & F1/Acc. & 0.430 & -     & -     & 0.920 & 0.421 & -     & - \\
    \midrule
    \midrule
    \multirow{2}[2]{*}{LLaMA3-8B-Instruct} & L.O.R. & 0.998 & 0.0   & 1.0   & 0.729 & 0.520 & 0.645 & 0.649 \\
          & F1/Acc. & 0.174 & 0.0   & 0.443 & 0.638 & 0.424 & 0.364 & 0.341 \\
    \midrule
    LLaMA3-8B-Instruct & L.O.R. & 1.0   & 0.685 & 1.0   & 1.0   & 1.0   & 1.0   & 0.947 \\
    (constrained with FSM) & F1/Acc. & 0.148 & 0.017 & 0.342 & 0.884 & 0.559 & 0.089 & 0.340 \\
    \midrule
    LLaMA3-8B-Instruct & L.O.R. & 1.0   & 0.970 & 1.0   & 1.0   & 0.998 & 1.0   & 0.995 \\
    (constrained with CFG) & F1/Acc. & 0.168 & 0.060 & 0.413 & 0.855 & 0.583 & 0.450 & 0.421 \\
    \midrule
    \multirow{2}[2]{*}{\archiname-8B-w.o.-ner} & L.O.R. & 0.979 & 0.999 & 1.0   & 0.983 & 0.838 & 0.968 & 0.961 \\
          & F1/Acc. & 0.719 & 0.653 & 0.515 & 0.935 & 0.525 & 0.466 & 0.635 \\
    \midrule
    \archiname-8B-w.o.-ner & L.O.R. & 1.0   & 0.012 & 1.0   & 1.0   & 1.0   & 1.0   & 0.835 \\
    (constrained with FSM) & F1/Acc. & 0.736 & 0.010 & 0.533 & 0.947 & 0.609 & 0.387 & 0.537 \\
    \midrule
    \archiname-8B-w.o.-ner & L.O.R. & 1.0   & 0.999 & 1.0   & 1.0   & 1.0   & 1.0   & 1.0 \\
    (constrained with CFG) & F1/Acc. & 0.706 & 0.650 & 0.515 & 0.948 & 0.617 & 0.475 & 0.652 \\
    \bottomrule
    \end{tabular}%
  \label{tab:exp-main}%
  }
\end{table*}%

%% file: tables/exp-data-weight.tex
\begin{table*}
  \centering
  \caption{Search experiments on the proportion of training datasets.}
  \small{
    \begin{tabular}{cc|l|ccccccc|c}
    \toprule
    \multicolumn{1}{l}{Task} & \multicolumn{1}{l|}{S2I} & Index & CNL03 & HOTEL & MED  & S10T8 & 20NEWS & DBP14 & RTE   & Avg \\
    \midrule
    \multirow{2}[2]{*}{20k} & \multirow{2}[2]{*}{0k} & L.O.R. & 0.718 & 1.0   & 0.774 & 0.990 & 0.999 & 0.984 & 1.0   & 0.924 \\
          &       & F1/Acc. & 0.712 & 0.529 & 0.456 & 0.692 & 0.658 & 0.926 & 0.769 & 0.677 \\
    \midrule
    \multirow{2}[2]{*}{17.5k} & \multirow{2}[2]{*}{2.5k} & L.O.R. & 0.826 & 1.0   & 0.742 & 0.993 & 0.999 & 0.970 & 1.0   & 0.933 \\
          &       & F1/Acc. & 0.753 & 0.545 & 0.450 & 0.704 & 0.657 & 0.918 & 0.791 & 0.688 \\
    \midrule
    \multirow{2}[2]{*}{15k} & \multirow{2}[2]{*}{5k} & L.O.R. & 0.871 & 0.993 & 0.807 & 0.960 & 0.988 & 0.986 & 1.0   & 0.943 \\
          &       & F1/Acc. & 0.765 & 0.567 & 0.457 & 0.634 & 0.644 & 0.920 & 0.765 & 0.679 \\
    \midrule
    \multirow{2}[2]{*}{10k} & \multirow{2}[2]{*}{10k} & L.O.R. & 0.851 & 0.993 & 0.774 & 0.994 & 0.983 & 0.988 & 1.0   & 0.940 \\
          &       & F1/Acc. & 0.751 & 0.497 & 0.458 & 0.643 & 0.646 & 0.924 & 0.791 & 0.673 \\
    \midrule
    \multirow{2}[2]{*}{5k} & \multirow{2}[2]{*}{15k} & L.O.R. & 0.918 & 0.989 & 0.774 & 0.978 & 1.0   & 0.981 & 1.0   & 0.949 \\
          &       & F1/Acc. & 0.739 & 0.516 & 0.425 & 0.589 & 0.578 & 0.925 & 0.812 & 0.655 \\
    \bottomrule
    \end{tabular}%
  \label{tab:exp-data-weight}%
  }
\end{table*}%

%% file: tables/exp-data-size.tex
\begin{table*}
  \centering
  \caption{Search experiments on the total training data volume.}
  \small{
    \begin{tabular}{c|l|ccccccc|c}
    \toprule
    \multicolumn{1}{l|}{Samples} & Index & CNL03 & HOTEL & MED  & S10T8 & 20NEWS & DBP14 & RTE   & Avg \\
    \midrule
    \multirow{2}[2]{*}{10k} & L.O.R. & 0.890 & 0.996 & 0.807 & 0.992 & 0.991 & 0.982 & 1.0   & 0.951 \\
          & F1/Acc. & 0.744 & 0.580 & 0.415 & 0.572 & 0.613 & 0.910 & 0.794 & 0.661 \\
    \midrule
    \multirow{2}[2]{*}{20k} & L.O.R. & 0.826 & 1.0   & 0.742 & 0.993 & 0.999 & 0.970 & 1.0   & 0.933 \\
          & F1/Acc. & 0.753 & 0.545 & 0.450 & 0.704 & 0.657 & 0.918 & 0.791 & 0.688 \\
    \midrule
    \multirow{2}[2]{*}{30k} & L.O.R. & 0.948 & 1.0   & 0.710 & 0.930 & 0.997 & 0.984 & 1.0   & 0.938 \\
          & F1/Acc. & 0.848 & 0.539 & 0.450 & 0.695 & 0.653 & 0.938 & 0.758 & 0.697 \\
    \midrule
    \multirow{2}[2]{*}{40k} & L.O.R. & 0.608 & 1.0   & 0.774 & 0.905 & 1.0   & 0.968 & 1.0   & 0.894 \\
          & F1/Acc. & 0.592 & 0.525 & 0.495 & 0.713 & 0.655 & 0.927 & 0.769 & 0.668 \\
    \bottomrule
    \end{tabular}%
  \label{tab:exp-data-size}%
  }
\end{table*}%

%% file: tables/schema-example-NER.tex
\begin{table}[ht]
\centering
\begin{minted}[fontsize=\scriptsize,frame=none,framesep=3pt,baselinestretch=0.1,bgcolor=bggray]{json}
{
    "type": "object",
    "properties": {
        "taskDesc": {"type": "string"},
        "entityTypes": {
            "type": "array",
            "items": {
                "type": "object",
                "properties": {
                    "name": {"type": "string"},
                    "description": {"type": "string"}
                },
                "required": ["name"]
            }
        },
        "outputFormat": {"type": "object"}
    },
    "required": ["taskDesc", "entityTypes", "outputFormat"]
}
\end{minted}
\caption{Schema for named entity recognition tasks.}
\end{table}

%% file: tables/schema-example-TC.tex
\begin{table}[ht]
\centering
\begin{minted}[fontsize=\scriptsize,frame=none,framesep=3pt,baselinestretch=0.1,bgcolor=bggray]{json}
{
    "type": "object",
    "properties": {
        "taskDesc": {"type": "string"},
        "labelSet": {
            "type": "array",
            "items": {
                "type": "object",
                "properties": {
                    "tag": {"type": "string"},
                    "description": {"type": "string"}
                },
                "required": ["tag"]
            }
        },
        "choiceType": {
            "type": "string",
            "enum": ["single", "multiple"]
        },
        "outputFormat": {"type": "object"}
    },
    "required": ["taskDesc", "labelSet", "choiceType", "outputFormat"]
}
\end{minted}
\caption{Schema for topic classification tasks.}
\end{table}

%% file: tables/schema-example-Translation.tex
\begin{table}[ht]
\centering
\begin{minted}[fontsize=\scriptsize,frame=none,framesep=3pt,baselinestretch=0.1,bgcolor=bggray]{json}
{
    "type": "object",
    "properties": {
        "taskDesc": {
            "type": "string"
        },
        "sourceLang": {
            "type": "string"
        },
        "targetLang": {
            "type": "string"
        },
        "outputFormat": {
            "type": "string"
        },
    },
    "required": ["taskDesc", "outputFormat"]
}
\end{minted}
\caption{Schema for machine translation tasks.}
\end{table}

%% file: tables/task-instantiation-NER.tex
\begin{table}[ht]
\centering
\begin{minted}[fontsize=\scriptsize,frame=none,framesep=3pt,baselinestretch=0.1,bgcolor=bggray]{json}
{
    "taskDesc": "Extract named entities from the text provided.",
    "entityTypes": [
        {"name": "person"}, {"name": "location"}, 
        {"name": "organization"}, {"name": "others"}
    ],
    "outputFormat": {
        "type": "array",
        "items": {
            "type": "object",
            "properties": {
                "name": {
                    "type": "string",
                    "description": "the entity name"
                },
                "entity_type": {
                    "type": "string",
                    "description": "entity type",
                    "enum": ["person", "organization", "location", "others"]
                }
            },
            "required": ["name", "entity_type"]
        }
    }
}
\end{minted}
\caption{An example of "Task Instance" in a named entity recognition task.}
\end{table}

%% file: tables/task-instantiation-TC.tex
\begin{table}[ht]
\centering
\begin{minted}[fontsize=\scriptsize,frame=none,framesep=3pt,baselinestretch=0.1,bgcolor=bggray]{json}
{
    "taskDesc": "Select a topic tag from the given options based on the article's content.",
    "labelSet": [
        {"tag": "World"}, {"tag": "Sports"}, 
        {"tag": "Business"}, {"tag": "Sci/Tech"}
    ],
    "choiceType": "single",
    "outputFormat": {
        "type": "object",
        "properties": {
            "tag": {
                "type": "string",
                "enum":["World", "Sports", "Business", "Sci/Tech"]
            }
        },
        "required": ["tag"]
    }
}
\end{minted}
\caption{An example of "Task Instance" in a topic classification task.}
\label{tab:ti-tc}%
\end{table}

%% file: tables/task-instantiation-Translation.tex
\begin{table}[ht]
\centering
\begin{minted}[fontsize=\scriptsize,frame=none,framesep=3pt,baselinestretch=0.1,bgcolor=bggray]{json}
{
    "taskDesc": "Translate the given text into target language.",
    "outputFormat": {
        "type": "object",
        "properties": {
            "translation": {
                "type": "string"
            }
        },
        "required": [
            "translation"
        ]
    }
}
\end{minted}
\caption{An example of "Task Instance" in a translation task.}
\label{tab:ti-translation}%
\end{table}